\documentclass{article} 
\usepackage{iclr2027_conference,times}
\iclrfinalcopy
\makeatother

\usepackage{amsmath,amsfonts,bm}

\def\eqref#1{equation~\ref{#1}}

\def\1{\bm{1}}

\DeclareMathAlphabet{\mathsfit}{\encodingdefault}{\sfdefault}{m}{sl}
\SetMathAlphabet{\mathsfit}{bold}{\encodingdefault}{\sfdefault}{bx}{n}

\usepackage{hyperref}
\usepackage{url}
\usepackage{natbib}
\setcitestyle{numbers,square}
\usepackage[utf8]{inputenc} 
\usepackage[T1]{fontenc}    
\usepackage{booktabs}       
\usepackage{amsfonts}       
\usepackage{nicefrac}       
\usepackage{microtype}      
\usepackage{xcolor}         
\usepackage{amsmath}
\usepackage{multirow}
\usepackage{graphicx}
\usepackage{pifont}
\usepackage{enumitem}
\newcommand{\cmark}{\ding{51}}
\newcommand{\xmark}{\ding{55}}

\title{M-Plan: Comprehensive Planning For LLM Agents with Memory Arrangement}

\author{Ziyang Liu$^1$, Jiakang Yuan$^1$, Jiayuan Fan$^2$, Tao Chen$^{1,3}$
\thanks{Corresponding author.} \\
$^1$College of Future Information Technology, Fudan University, Shanghai, China \\
$^2$College of Intelligent Robotics and Advanced Manufacturing, Fudan University, Shanghai, China \\
$^3$Shanghai Innovation Institute, Shanghai, China \\
\texttt{ziyangliu25@m.fudan.edu.cn~~eetchen@fudan.edu.cn}
}

\begin{document}

\maketitle

\begin{abstract}
Large language model (LLM)-based agents have recently demonstrated outstanding performance in single-turn tasks, yet they continue to struggle with long-horizon tasks, particularly in environments without prior training. A key limitation of single LLMs is their difficulty in processing long contexts, which stems from weaknesses in continuous decision-making and reasoning capabilities. Although existing multi-agent systems (MAS) equipped with long-term memory bases can partially address this gap, they mostly rely on knowledge reuse or summarization—an approach that is often constrained by directly retrieval without any adaptation for the current task, while the retrieved memory may contain much history noises waiting to be preprocessed. In contrast, MAS characterized by task-specific guidance often lack long-term memory specifically designed for the guidance. 
To address these challenges, we propose M-Plan, a closed-loop MAS framework that integrates long-term and short-term memory. By combining global-local and past-current guidance, M-Plan enhances agents' performance on multi-turn tasks while maintaining a plug-and-play design. During task execution, agents receive task-specific milestones and insights, and are provided with local hints only when confusion arises, which form global-local guidance together. These guidance generation and utilization construct the short-term memory as well. After task completion, relevant information about the current task is optimized or reorganized before being sent to the long-term memory storage for future retrieval. The designments mentioned above build a comprehensive framework different from other individual long-term or short-term memory works. Extensive experiments on multi‑turn tasks including ALFWorld, Webshop, and PDDL across diverse models demonstrate that M‑Plan outperforms multiple baseline methods in both accuracy and efficiency.
\end{abstract}

\section{Introduction}

Large language models (LLMs) have been widely applied across numerous domains, and extensive research has demonstrated their strong capability to solve a diverse range of tasks~\cite{xi2023risepotentiallargelanguage}. With the increasing continuity and complexity of real-world tasks in daily human activities, LLMs are required to tackle multi-turn, long-horizon tasks more frequently. Such tasks demand sustained decision-making and reasoning across multiple turns, while also requiring models to maintain and process long contexts. However, these demands remain challenging for existing LLMs, which, despite their remarkable performance on single-turn instructions, often struggle in multi-turn scenarios~\cite{laban2025llmslostmultiturnconversation,li2026singleturnsurveymultiturninteractions}. Common issues observed in LLM-based agents include forgetting information from intermediate turns, overemphasizing initial and final turns, which indicate that a single LLM is insufficient to handle increasingly complex contexts.

To address these challenges, researchers have proposed two major approaches for adapting LLM-based agents. The first exploits training to optimize agents on downstream tasks \cite{forouzandeh2025learninghierarchicalproceduralmemory,wang2026rlanythingforgeenvironmentpolicy,li2026dynamicmixprecisionrouting,xi2025agentgymrltrainingllmagents,peng2026hiperhierarchicalreinforcementlearning,zhou2025mementofinetuningllmagents,ren2026magicagentgeneralizedagentplanning}, primarily including Supervised Fine-Tuning (SFT) and Reinforcement Learning (RL), typically with human-annotated training sets, which are sometimes hard to get. The second adopts training-free methods, which typically employ a specialized multi-agent system (MAS) to process long contexts. Most existing training-free approaches integrate MAS with long term memory mechanisms \cite{zhang2025gmemorytracinghierarchicalmemory,wang2025mirixmultiagentmemoryllmbased,zhang2026memskilllearningevolvingmemory,ouyang2026reasoningbankscalingagentselfevolving}, usually incurring lower computational costs with greater flexibility and interpretability. For example, through setting specific agents to examine the task execution result, process successful and failed trajectories, ReasoningBank \cite{ouyang2026reasoningbankscalingagentselfevolving} can learn from historical experiences to adapt to other tasks, and construct a visual external memory base, which is more interpretable than training.
As for short-term memory, characterized by a combination of global and step-wise guidance \cite{zhou2024tradenhancingllmagents,hu2024hiagenthierarchicalworkingmemory, li2025hiplanhierarchicalplanningllmbased}, those works are able to build a overall route, possibly with local hints, helping the action agent make decision more wisely.

Despite remarkable progress in MAS-based approaches, the aforementioned methods still suffer from several limitations. For long-term memory, typical approaches involve reusing historical experience or summarizing general knowledge, but these methods struggle to ensure generalization, and long historical trajectories may introduce noise and instability. For short-term memory, step-wise generation inevitably incurs substantial token consumption, whereas relying solely on global guidance often fails to provide complete task instructions, since high-level blueprints or milestones lack the necessary details and action-related knowledge required for execution. In addition, existing MAS-based planning works generally lack memory association, which weakens their self-evolution capabilities and limits their applicability to real-world scenarios. 

\begin{figure}[tbp]
   \centering
   \includegraphics[width=1.0\textwidth]{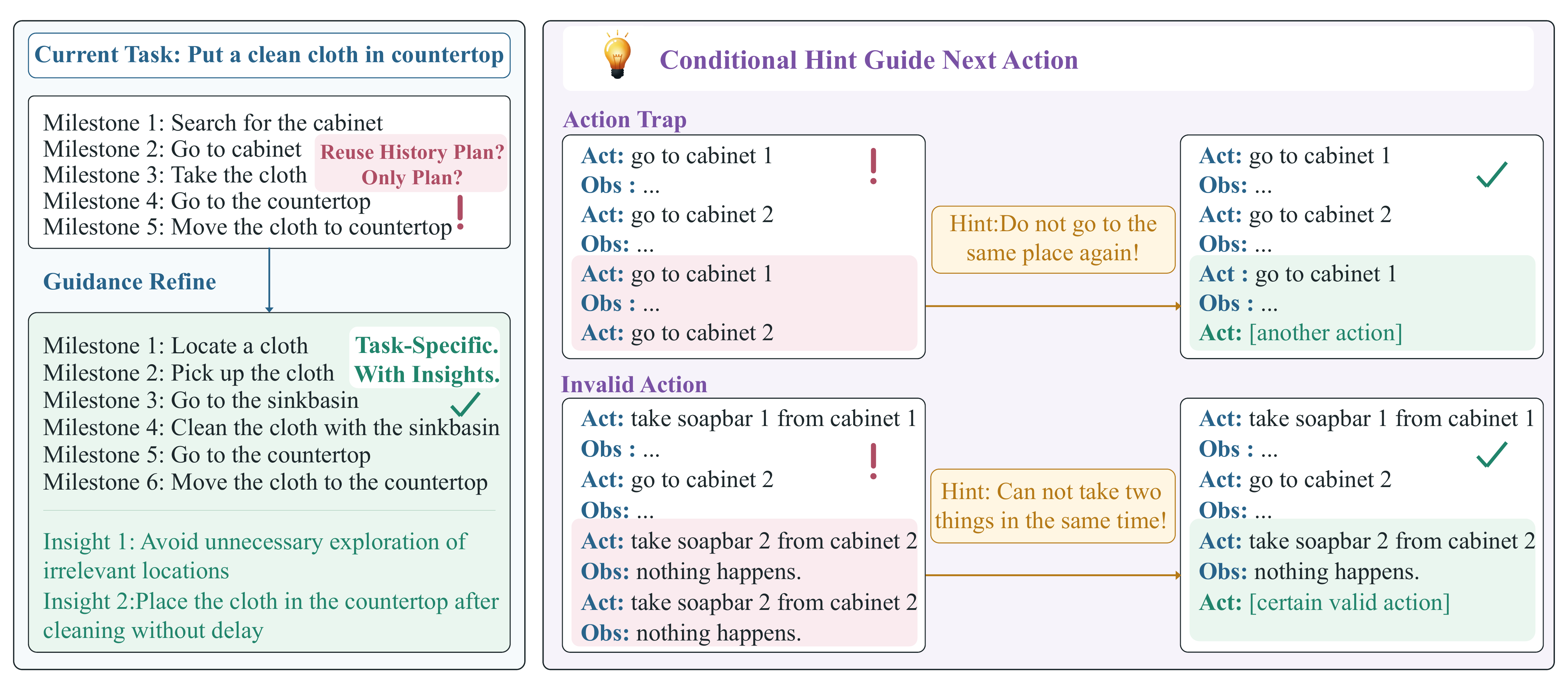}
   \caption{The task-specific guidance consists of milestones and insights: milestones guide the action agent to follow a global task-solving path, while insights provide detailed considerations and important task-relevant knowledge; conditional hint when meeting action trap (invalid action). The hint will help correct the action in the next step.}
   \label{fig:1}
\end{figure}

To address these limitations, we propose M-Plan, a long and short-term memory MAS framework. The short-term memory integrates comprehensive task-specific global guidance, as shown in the left part of Fig.\ref{fig:1} and local hints in the right part. With task-specific milestones and insights, our short-term memory can digest the retrieved information into task-specific guidance by refusing simply retrieval and adding insights to complement details, solving reuse problems in individual long-term memory. Not like the former step-wise guidance work, considering that action traps and invalid actions remain prevalent in most multi-turn tasks, our conditionally triggered hints, can reduce token consumption while helping agents correct their actions in subsequent steps. However, a short-term memory usually lacks a great partner in similar work, and our long‑term memory complements short‑term memory for this purpose. Precisely, to better adapt agents to new tasks, M-Plan maintains a long-term memory base for guidance, helping task-specific short-term memory further enhances the effectiveness of long-term memory during the current task. To improve the quality of information stored in the memory base and connect the short- and long-term memory, we further design an \textsc{Optimizer} module to refine completed task trajectories before they are stored as long-term memory, helping the short- and long-term memory achieve better complementarity.

We evaluate M-Plan on three commonly used multi-turn benchmarks: ALFWorld\cite{shridhar2021alfworldaligningtextembodied}, Webshop\cite{yao2023webshopscalablerealworldweb} and PDDL\cite{ma2024agentboardanalyticalevaluationboard}, which are filled with long-horizon and context-rich tasks. Experiments on two models of different scales demonstrate that M-Plan outperforms classic and typical baselines in terms of effectiveness (accuracy and reward). To assess efficiency, we report the average steps, redefined steps and compare valid token consumption of each successful task. Additionally, ablation studies verify the value of each designed module in enhancing performance. 

The contributions of this paper can be summarized as follows:

\begin{enumerate}
    \item We propose M-Plan, a MAS framework that integrates long-term memory storage, short-term memory planning mechanisms, and global-local guidance. This framework eliminates the need for reusing abstract knowledge, maintains the generation of valuable guidance through long-term memory, avoids uncertainty, and controls token costs via conditional local guidance. This comprehensive closed-loop framework is effective across different models and can be theoretically extended to various multi-turn tasks.
    \item We construct a closed-loop and complete system. In M-Plan, milestones and insights form a complete guidance system, with conditional local hints providing step-wise assistance when agents encounter confusion, with long term memory offering information from optimizer as generation references, which can abandon traditional reuse-based approaches for guiding agent-environment interactions.
    \item We conduct experiments on challenging benchmarks, and by leveraging task-specific information, M-Plan utilizes tokens more efficiently while outperforming all baseline methods using different base models.
\end{enumerate}

\begin{figure}[tbp]
   \centering
   \includegraphics[width=1.0\textwidth]{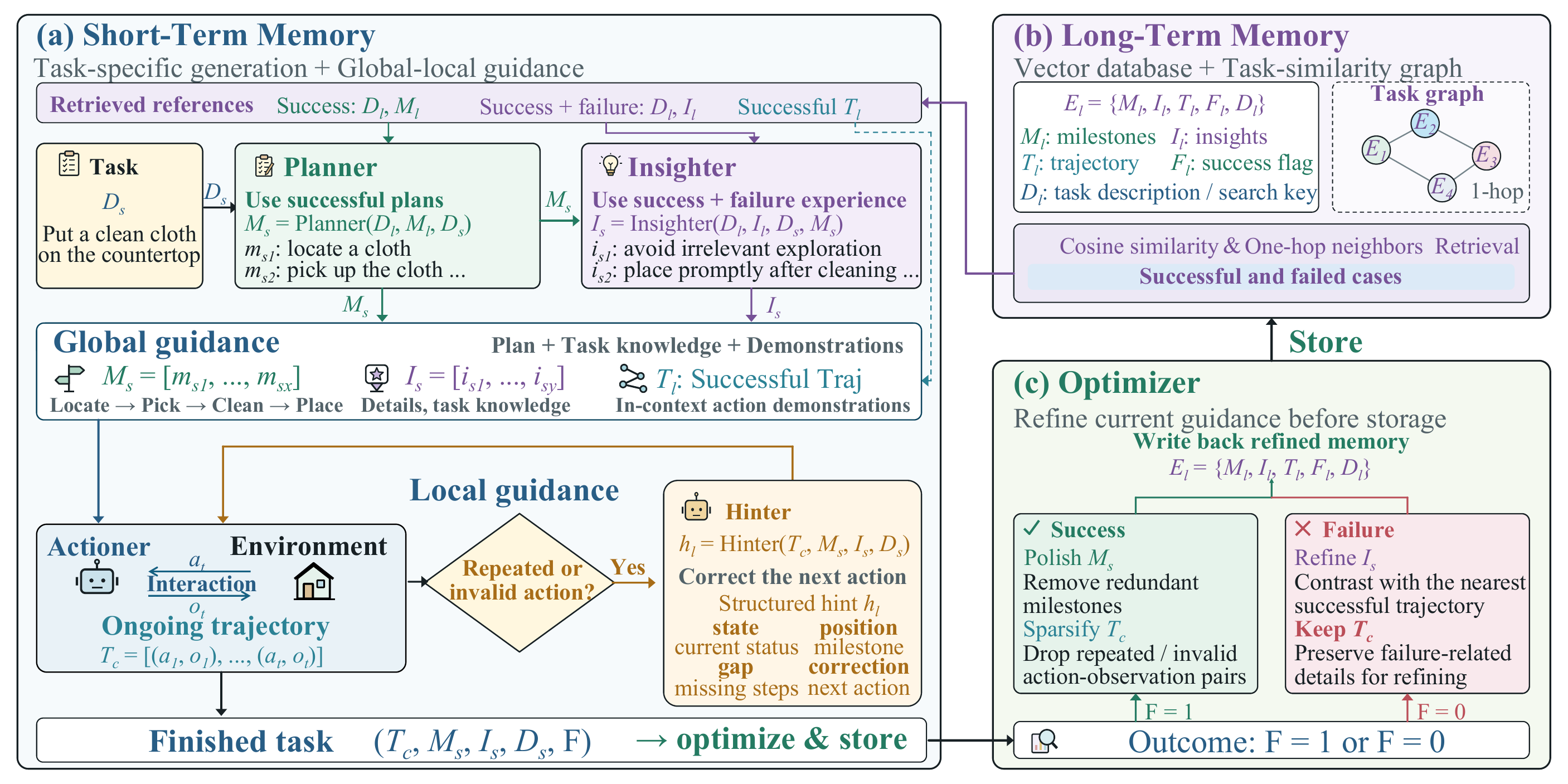}
   \caption{The whole framework of M-Plan, containing long short-term memory, global-local guidance (milestones, insights, hints), and an optimizer module to form the closed-loop. More details can be seen in Section \ref{method}.}
   \label{fig:2}
\end{figure}

\section{Related Work}

\subsection{Long Term Memory}

Existing long-term memory works can be broadly categorized into parametric or non-parametric approaches. Parametric methods typically rely on large-scale trajectories or other pre-collected information to train models through reinforcement learning (RL) or supervised fine-tuning (SFT), thereby encoding long-term memory into model parameters~\cite{li2025memosmemoryosai,Yang_2024,liu2025memversemultimodalmemorylifelong}.
For example, MEMOS~\cite{li2025memosmemoryosai} proposes a large and complex memory operating system with parametric memory storage. Memento~\cite{zhou2025mementofinetuningllmagents} further uses RL to improve retrieval quality.
However, parametric memory is still ultimately derived from non-parametric data, such as textual knowledge or trajectories. Although training can improve agents' performance on downstream tasks, it introduces substantial computational costs and usually requires complex data preparation in advance.

To support plug-and-play and convenient deployment, lots of works design non-parametric memory~\cite{zhang2025gmemorytracinghierarchicalmemory,ouyang2026reasoningbankscalingagentselfevolving,zhang2026memskilllearningevolvingmemory,tang2025chemagentselfupdatinglibrarylarge,zhong2023memorybankenhancinglargelanguage,park2023generativeagentsinteractivesimulacra}. A typical method~\cite{wang2024agentworkflowmemory,liang2026learningremembermetacognitivemanagement,packer2024memgptllmsoperatingsystems,LIANG2025130470} is to construct memory bases using vectors or textual representations for semantic retrieval. A representative paradigm is to store extracted, summarized, or general knowledge and reuse it to guide agents in action execution. For example, work \cite{zhang2025gmemorytracinghierarchicalmemory} sets a series of agents refining, editing, and integrating general insights to be reused in future tasks, and also extracts the key steps similar to the plan of the history trajectories as in-context content. However, these methods often struggle to extract truly transferable knowledge and usually lack task-specific adaptation. Directly using retrieved memories may also introduce semantic noise from other tasks. Although historical trajectories can serve as useful examples, they are not always reliable in multi-step decision-making scenarios. To improve robustness and task-specific adaptation, M-Plan introduces a short-term memory mechanism on top of long-term memory. We further design an \textit{Optimizer} module to compress and refine completed task experiences, forming a closed-loop process that feeds successful experiences and failure lessons back into long-term memory. In this way, retrieval and feedback integrate long-term and short-term memory into a unified system.

\subsection{Short Term Memory}

Agent-based short-term memory mechanisms for planning and acting have shown significant potential in multi-turn interactive tasks. Existing methods can be roughly divided into two categories. The first category relies on global guidance \cite{sun2023pearlpromptinglargelanguage}, step-wise guidance \cite{erdogan2025planandactimprovingplanningagents,nguyen2024steptimelanguageagents}, or their combination. For example, HiAgent~\cite{hu2024hiagenthierarchicalworkingmemory} segments the working trajectory into subgoals to organize the context, enabling partial local compression and guidance. HiPlan~\cite{li2025hiplanhierarchicalplanningllmbased} introduces a planning system equipped with global milestones and step-wise hints.
Another common approach is to insert reasoning traces into the action trajectory, with ReAct~\cite{yao2023reactsynergizingreasoningacting} as a typical example, followed by related studies~\cite{zhou2024tradenhancingllmagents,sarukkai2026inferencetimedistillationcostefficientagents}. However, these methods overlook the fact that plans alone cannot always provide sufficient guidance for task completion. Moreover, step-wise generation or judgment can introduce substantial time and token costs. The second category adopts retrieval-augmented planning, such as~\cite{zhou2024tradenhancingllmagents,li2025hiplanhierarchicalplanningllmbased}. Although these methods are convenient and effective, they often rely on few-shot examples like full trajectories or other static memory. To provide deeper and more comprehensive guidance during task execution, M-Plan introduces an \textit{Insighter} into the short-term memory module. We further design a dynamic external memory base for task-level retrieval, enabling short-term memory to evolve across different agents.

\subsection{Long Short-Term Memory}

Several existing works have introduced systems that combine long-term and short-term memory \cite{liu2025memversemultimodalmemorylifelong,wen2026aligningprogressfeasibilityneurosymbolic,sun2026kernelskillmultiagentframeworkgpu}. For example, MemVerse~\cite{liu2025memversemultimodalmemorylifelong} adopts a sliding window to process short-range dialogue information and uses a relational graph to store task information, including episodic, core, and semantic memories in textual form. It further employs a supervised fine-tuned model to transform graph-based memories into model weights. Another work~\cite{wen2026aligningprogressfeasibilityneurosymbolic} constructs process and feasibility memories to organize knowledge from successful and failed trajectories, while also monitoring task progress locally. However, these methods typically require retrieval or complex memory modification at each step, which may slow down execution and substantially increase computational cost. In contrast, our system avoids frequent step-wise retrieval or examination while maintaining strong performance and further establishes a closed-loop optimization mechanism through the interaction between long-term and short-term memory.

\section{Methodology}
\label{method}

\subsection{Overview}

As shown in Figure \ref{fig:2}, from the function view, M-Plan can be separated into three modules: long-term memory storage, short-term memory generation, and optimizer. Following M-Plan's workflow, there will be a closed-loop in the whole, meaning that results made by short-term memory can influence it in a future turn, and information derived from long-term memory can enlarge the base after finishing the current task. Overall, with MAS design, whose prompts are presented in Appendix. \ref{A}, M-Plan can utilize the following two kinds of information as guidance to help action agents execute the task:

\begin{enumerate}
    \item Global guidance: A generated milestones list  $M_g=[m_1,m_2,m_3,...,m_x]$,  playing the role of  planning route, where each $m_x$ represents one key subgoal need to achieve. A generated insights list $I_g=[i_1,i_2,i_3,...,i_y]$,  where each $i_y$ represents one key insight as noticeable reference. Historically successful trajectories $T_g$ from memory base provide demonstrations and experiences like the human recalling when meeting similar tasks. 
    \item Local guidance: A structural hint $h_l$ can be included to the context to remind the action agent to get rid of the current problems, performing as a local guide. However, only under the conditions where the action agent is confused, the hint will be generated and exploited. This strategy can be likened to a wise, deep-thinking pattern with an external brain assistant, and also conforms to human thinking patterns, in which people will think more deeply to tackle traps or trouble. 
\end{enumerate}

\subsection{Long-term Memory Storage}

We first introduce the long-term memory storage module, able to break the limitation of single short-term memory in most work. This module equips agents with historical task information as references, thereby enabling more effective task-specific generation. Notably, the memory base does not rely on any pre-collected data or demonstrations. Instead, it starts from scratch and continuously absorbs external information from short-term memory during task execution. Specifically, Each memory item $E_l$ is defined as follows: 
\begin{equation}
    E_l=\{M_l,I_l,T_l,F_l,D_l\},
\end{equation}
where $M_l$ denotes task milestones, $I_l$ represents task insights, $T_l$ records the execution trajectory, $F_l$ is a Boolean flag indicating whether the trajectory succeeds or fails, and $D_l$ is the task description, which serves as the key for semantic search. These memory items are stored in a vector database. In addition, following the graph-based memory organization strategy in \cite{zhang2025gmemorytracinghierarchicalmemory}, we model the relationships among memory items in graph form, where edges are added between similar tasks to facilitate faster retrieval and better structural organization.

To better utilize long-term memory, the retrieval process can be formulated as follows:
\begin{equation}
\begin{aligned}
    &\{E_1, E_2, ... E_j\}=Ret(Q), ES=\{E_1, E_2, ... E_k\} \\
    &\{E_1, E_2, ..., E_j, ... , E_k\}=Extend(ES),
\end{aligned}
\end{equation}
where $Ret(Q)$ denotes the retrieval function triggered by the task query $Q$, which retrieves $j$ relevant memory items from the memory base according to semantic cosine similarity. The retrieved set is denoted as $ES$. To further improve diversity and robustness, we apply a one-hop graph expansion strategy to enlarge the retrieval scope and obtain the final top-$k$ memory items. Specifically, for each item in $ES$, we search its directly connected one-hop neighbors in the memory graph and append them to the original retrieved set, beyond those retrieved by semantic similarity.

\subsection{Short-term Memory Generation}

However, current work relying solely on long-term memory may be insufficient for task-specific enhancement because historical experiences mainly provide coarse references from past tasks and cannot fully capture the dynamic states, and task-specific constraints of the current execution process. Moreover, most existing short-term planning works generate only coarse-grained plans, lacking sufficient details. And step-wise guidance at every step can also introduce substantial computational cost. To address these limitations, we introduced a multi-agent based short-term memory, including a $Planner$, an $Insighter$, a $Hinter$, and an $Actioner$. Specifically, we first retrieve two successful cases and one failed case from long-term memory. The two success trajectories serve as fundamental in-context demonstrations. The $Planner$ receives the description $D_l$ and milestones $M_l$ of past successful trajectories for positive reference, together with the current task description $D_s$, and generates a new milestone list for the current task:
\begin{equation}
    \{m_{s1}, m_{s2}, m_{s3}, ... , m_{sx}\}=Planner(D_l,M_l,D_s).
\end{equation}
The $Insighter$ receives the milestone list $M_s=\{m_{s1}, m_{s2}, m_{s3}, ... , m_{sx}\}$, adding more related details to enrich the content of the blueprint, and takes the task description $D_l$ from both successful and failed cases, along with the current task description $D_s$, the current task milestones and their corresponding insight list $I_l$, to identify the information that may be valuable for the current task. It then generates a new insight list:
\begin{equation}
    \{i_{s1}, i_{s2}, i_{s3}, ... , i_{sy}\}=Insighter(D_l,I_l,D_s,M_s).
\end{equation}
Then, the $Hinter$ receives $D_s$, the ongoing trajectory $T_c$, the global milestone list $M_s$ and insight list $I_s= \{i_{s1}, i_{s2}, i_{s3}, ... , i_{sy}\}$ as input. The hint-generation process is formulated as follows: 
\begin{equation}
    \{state,position, gap, correction\}=Hinter(T_c,M_s,I_s,D_s).
\end{equation}

To reduce computational cost, we adopt a conditional step-wise hinting strategy. Instead of guiding every step, the $Hinter$ is activated only when the $Actioner$ falls into an action trap or performs an invalid action according to the environment feedback. More detailed rules of the $Hinter$ are expanded on Appendix. \ref{D}.
The generated hint consists of four components: $state$, which describes the agent's current state, such as objects it has already collected; $position$, which indicates the current milestone to be achieved; $gap$, which provides advice on how to complete the current milestone; and $correction$, which suggests an action correction for the next step.

The $Actioner$ is responsible for selecting and executing actions based on the retrieved successful trajectories, current observation, task description, generated milestones, insights, and conditional hints.

\subsection{Optimizer}

Apart from long and short-term memory, there remains a problem of how to process and exploit the information from finished tasks, which is also rarely explored in current work. When $Planner$ or $Insighter$ directly receives unrefined information, their performance may be hindered, especially when the long-term memory is still in a raw stage. To address this issue, we design an $Optimizer$ module to reflect on finished task trajectories, including both successful and failed ones, and to form a closed-loop optimization process. Specifically, the $Optimizer$ works by refining the current task-specific generation, rather than entirely restarting summarization or extraction. The $Optimizer$ can be divided into three small parts: milestone list polishing for $M_s$, insight list complement for $I_s$, and trajectory sparsification for the entire task trajectory $Traj$.

In our former exploration phase, we found that a successfully executed multi-turn task may still have an imperfect $M_s$, which can be polished by analyzing the whole trajectory. When the task successfully finished, the $optimizer$ is more likely to discover redundant milestones during execution. However, it is harder for $optimizer$ to polish $M_s$ when the task fails, because the failure is mostly caused by not following $M_s$ well. In other words, $M_s$ is usually correct, though it may not be perfect, which is why we only polish it when the task succeeds. In contrast, for $I_s$, it is harder to extract more valuable information from successful tasks, but notable points can often be obtained from failed ones. Moreover, given that refining $I_s$ solely from the current failed trajectory remains challenging, we also input the most similar retrieved successful trajectory to facilitate this refinement by contrasting failed and successful rollouts. The sparsification of $Traj$ is much simpler, since it only removes repeated or invalid action-observation pairs in the trajectory. Note that sparsification is only applied to successful trajectories, since failed ones need to keep failure-related details. Overall, the outputs of the three parts are derived from the short-term memory mechanism above and will be sent to long-term memory storage as references for short-term memory generation in the next task, and then form a complete closed loop between long-term and short-term memory.

\section{Experiment}

We evaluate M-Plan on three widely test multi-turn benchmarks:ALFWorld \cite{shridhar2021alfworldaligningtextembodied}, Webshop \cite{yao2023webshopscalablerealworldweb}, PDDL \cite{ma2024agentboardanalyticalevaluationboard}, and take G-Memory \cite{zhang2025gmemorytracinghierarchicalmemory}, ReasoningBank \cite{ouyang2026reasoningbankscalingagentselfevolving} and Hiplan \cite{li2025hiplanhierarchicalplanningllmbased} as baselines. The detailed benchmark and baseline introduction can be seen in Appendix. \ref{C}.

\subsection{Metrics}
\label{metric}

\textbf{Effectiveness}: For ALFWorld~\cite{shridhar2021alfworldaligningtextembodied}, we report the success rate (\texttt{SR.}) to indicate the proportion of tasks completed successfully. For Webshop~\cite{yao2023webshopscalablerealworldweb} and PDDL~\cite{ma2024agentboardanalyticalevaluationboard}, in addition to \texttt{SR.}, we report the average reward to reflect the degree of attribution alignment (\textit{i.e.}, how well the purchased objects satisfy the target requirements). 

\textbf{Efficiency}: We report the average number of steps over all tasks (\texttt{AS.}) for ALFWorld~\cite{shridhar2021alfworldaligningtextembodied} and PDDL~\cite{ma2024agentboardanalyticalevaluationboard} to directly compare the action efficiency of different agents. In addition, we separately report the average number of steps for successful tasks (\texttt{ASS.}), which reflects efficiency under a fixed success rate. However, a common phenomenon in these benchmarks is that a lower \texttt{SR.} may lead to a lower \texttt{ASS.}, because the agent may only succeed on easier tasks that require fewer steps. Therefore, for a fairer comparison, we define \texttt{Meta-Step}, computed as $\texttt{ASS./SR.}$, as a more comprehensive efficiency metric. For Webshop~\cite{yao2023webshopscalablerealworldweb}, unlike ALFWorld, there is no fixed number of steps for failed tasks, and most failures result from incorrect purchases made in early steps. Thus, a smaller \texttt{AS.} does not necessarily indicate better efficiency. Therefore, for Webshop, we mainly focus on \texttt{Meta-Step}.

Another important dimension of efficiency is token usage. Given that a certain number of tokens are consumed to complete a task, both gains and losses should ideally be considered. However, the loss caused by task failure is difficult to quantify in these benchmarks. Therefore, we adopt $Ft$ as a more robust and interpretable measure of pure profit, defined as:
\begin{equation}
    Ft=\frac{2}{\pi}*\arctan(SR/FR),\quad
    Ct=\frac{At}{Ft}.
\end{equation} 
where $At$ denotes the average number of tokens consumed across all tasks and $FR$ denotes the failure rate, and both $SR$ and $FR$ are normalized. When $At$ is fixed, a lower $Ct$ and a higher $Ft$ indicate that the same token budget is expected to yield a greater advantage of success over failure. There will be detailed discussion about how we get this equation in the Appendix. \ref{B}. Theoretically, $At$ is positively correlated with \texttt{AS} under a fixed framework when each action consumes a similar number of tokens. However, we observed that in Webshop, certain actions such as examining product information incur substantially higher token overhead than simple click operations, breaking this balance. We therefore discard this metric for the Webshop environment to eliminate action-level unfairness.

\subsection{Main Results}

\begin{table}[t]
  \caption{Results on ALFWorld, Webshop and PDDL. We report Success Rate (\texttt{SR.}) on all three benchmarks and Average Reward ($AD$) on Webshop and PDDL. Details about metrics can be seen in Sec.~\ref{metric}. All \texttt{SR.} and $AD$ values are reported in percentage (\%).}
  \label{effectiveness}
  \centering
  \resizebox{1.0\linewidth}{!}{
  \begin{tabular}{lccccccc}
    \toprule
\multirow{2}{*}{Models} & \multirow{2}{*}{Methods}  & ALFWorld & \multicolumn{2}{c}{Webshop} & \multicolumn{2}{c}{PDDL} \\
\cmidrule(lr){3-3}  \cmidrule(lr){4-5} \cmidrule(lr){6-7}
& & \texttt{SR.} & \texttt{SR.} & $AD$ & \texttt{SR.} & $AD$\\
\midrule
\multirow{4}{*}{Qwen3-32B}
& ReasoningBank       & 69.4\textpm0.0 & 26.8\textpm1.2 & 53.5\textpm2.5 & 23.9\textpm1.0 & 47.3\textpm1.4 \\
& G-Memory            & 85.0\textpm1.3 & 46.8\textpm0.8 & 70.0\textpm0.3 & 25.6\textpm2.0 & 47.0\textpm2.8\\
& Hiplan              & 94.3\textpm0.9 & 39.5\textpm0.0 & 52.5\textpm0.0 & 22.2\textpm0.9 & 39.2\textpm0.7\\
& \textbf{M-Plan}     & \textbf{99.3\textpm0.0} & \textbf{56.5\textpm0.0} & \textbf{74.7\textpm0.0} & \textbf{52.8\textpm0.9} & \textbf{74.3\textpm1.3}\\
\midrule
\multirow{4}{*}{Qwen3-14B}
& ReasoningBank       & 65.7\textpm0.0 & 28.2\textpm0.6 & 60.7\textpm1.7 & 17.2\textpm0.9 & 37.9\textpm0.3\\
& G-Memory            & 84.6\textpm0.5 & 57.3\textpm1.2 & 79.8\textpm3.1 & 24.4\textpm2.0 & 41.8\textpm5.2\\
& Hiplan              & 87.8\textpm1.7 & 32.0\textpm0.0 & 44.8\textpm0.0 & 15.0\textpm1.7 & 25.5\textpm0.8\\
& \textbf{M-Plan}     & \textbf{90.8\textpm0.4} & \textbf{68.2\textpm0.3} & \textbf{82.6\textpm0.1} & \textbf{46.1\textpm1.9} & \textbf{68.9\textpm0.5}\\
\bottomrule
  \end{tabular}
  }
\end{table}

\begin{table}[t]
  \caption{Comparison of average steps on ALFWorld, Webshop and PDDL. We report \texttt{AS.}, \texttt{ASS.}, and \texttt{Meta-Step}, while \texttt{AS.} is not considered for Webshop due to its specific characteristics, as discussed in Sec.~\ref{metric}.}
  \label{steps}
  \centering
  \resizebox{1.0\linewidth}{!}{
  \begin{tabular}{l c c c c c c c c c}
    \toprule
    \multirow{2}{*}{Models} & \multirow{2}{*}{Methods} & \multicolumn{3}{c}{ALFWorld} & \multicolumn{2}{c}{Webshop} & \multicolumn{3}{c}{PDDL}\\
    \cmidrule(lr){3-5} \cmidrule(lr){6-7} \cmidrule(lr){8-10}
    & & \texttt{AS.} & \texttt{ASS.} & \texttt{Meta-Step} & \texttt{ASS.} & \texttt{Meta-Step} & \texttt{AS.} & \texttt{ASS.} & \texttt{Meta-Step}  \\
    \cmidrule(lr){1-10}
    \multirow{4}{*}{Qwen3-32B}
    & ReasoningBank       & 24.99 & 13.96 & 20.12& 6.14 & 22.90&26.61 &15.76 & 65.90\\
    & G-Memory            & 24.93& 20.53& 24.14& \textbf{5.10} & 10.90&26.95&18.16& 71.73 \\
    & Hiplan              & 18.38 & 16.46& 17.46& 8.09& 20.48&26.33&\textbf{13.48} & 60.70\\
    & \textbf{M-Plan}     & \textbf{12.48} & \textbf{12.20} & \textbf{12.28} & 5.35& \textbf{9.47} &\textbf{22.30} & 15.42 & \textbf{29.22} \\
    \cmidrule(lr){1-10}
    \multirow{4}{*}{Qwen3-14B}
    & ReasoningBank       & 26.30& 13.91& 21.17& 5.90 & 20.93&28.16 &19.33& 112.42\\
    & G-Memory            & 18.66 & 12.94 & 15.30& \textbf{5.13} & 8.96 &26.97 &17.54 & 71.86\\
    & Hiplan              & 20.40 & 16.28& 18.55 & 8.05 & 25.16 &27.56 &\textbf{13.71} &92.18 \\
    & \textbf{M-Plan}     & \textbf{14.63} & \textbf{11.05} & \textbf{12.17} & 5.48 & \textbf{8.03} &\textbf{23.33} &15.54 & \textbf{33.75}\\
    \bottomrule
  \end{tabular}
  }
\end{table}

We present our main results with performing each experiment three times to evaluate the possible variances, in Table~\ref{effectiveness}, Table~\ref{steps}, and Table~\ref{token}. To improve the experience of reading, all results of step and token with standard variance are presented in Appendix.~\ref{E}. These results compare both effectiveness and efficiency, showing that M-Plan consistently outperforms all baselines in both aspects. We provide a detailed analysis below.

\textbf{Effectiveness}: Table~\ref{effectiveness} summarizes the effectiveness results in terms of success rate (\texttt{SR.}) and average reward (AD). M-Plan achieves state-of-the-art performance with both Qwen3-32B and Qwen3-14B backbones.
On ALFWorld, M-Plan attains the highest \texttt{SR.} of 
99.3±0.0
 using Qwen3-32B, surpassing Hiplan (the second-best baseline) and substantially outperforming the memory-reuse method ReasoningBank. M-Plan maintains strong performance across diverse task categories without obvious failure modes on specific task types.
On Webshop, M-Plan exhibits an even larger performance gap compared with baselines. With Qwen3-32B, M-Plan improves \texttt{SR.} by roughly 20\% and average reward by around 7\% relative to G-Memory. When switching to the smaller Qwen3-14B backbone, the performance advantage over G-Memory narrows on ALFWorld, which we attribute to weaker capability of the smaller LLM to generate high-quality task-specific guidance. Even so, M-Plan consistently surpasses memory-reuse baselines such as G-Memory, verifying the benefits of our task-specific short-term memory design and the dedicated optimizer modules. Compared with the step-wise planning baseline Hiplan, the gains demonstrate that M-Plan can yield superior success rates without injecting hints at every single action step.
On the more challenging PDDL benchmark, M-Plan achieves a clear margin over all baselines: with Qwen3-32B, its \texttt{SR.} reaches 
52.8±0.9, more than doubling the success rate of ReasoningBank and G-Memory, suggesting that even in complex scenarios, M-plan can achieve great performance.

\textbf{Step Efficiency}: We analyze action efficiency using step-based metrics in Table~\ref{steps}.
On ALFWorld, M-Plan achieves the lowest \texttt{AS.} and \texttt{Meta-Step} under both model sizes. For Qwen3-32B, M-Plan reduces \texttt{Meta-Step} to 
12.28
, around 30\% lower than Hiplan. This indicates that M-Plan’s \textit{Actioner} leverages memory and guidance effectively to complete tasks with fewer actions while maintaining high success.
For Webshop, where \texttt{AS.} is not a reliable metric, we focus on \texttt{Meta-Step}. M-Plan achieves the smallest \texttt{Meta-Step} in both model settings. With Qwen3-32B, M-Plan cuts the redefined step overhead by over 50\% relative to Hiplan and around 10\% compared with G-Memory.
On PDDL, M-Plan also delivers the lowest \texttt{Meta-Step}, showing that the planning and memory mechanism reduces redundant actions even for complex planning tasks. These step metrics collectively show that M-Plan improves action efficiency without sacrificing task success.

\textbf{Token Efficiency}: We further evaluate token-level efficiency with $At$, $Ft$, and $Ct$ in Table~\ref{token}. $Ft$ measures the pure success profit, while $Ct$ represents the normalized token cost per unit profit: a smaller $Ct$ means the agent gains stronger success-failure advantage under the same token budget.
Across ALFWorld and PDDL, M-Plan yields substantially lower $Ct$ than all baselines. On ALFWorld with Qwen3-32B, M-Plan’s $Ct$ is nearly 70\% lower than Hiplan, demonstrating that our framework extracts higher task profit at a much cheaper token cost. The step-wise baseline Hiplan suffers particularly high token consumption, especially on PDDL, which implies that per-step local guidance incurs heavy token overhead that may offset its planning benefits. In contrast, M-Plan’s selective hinting and memory retrieval strategy reduce unnecessary LLM invocations, achieving better success with more economical token usage.

\begin{table}
  \caption{Token-efficiency comparison. We report $At$, $Ft$ and $Ct$ (see in Sec.~\ref{metric}), mainly focusing on the comparison of $Ct$. Webshop is not discussed for the fairness in Sec.~\ref{metric}.}
  \label{token}
  \centering
  \resizebox{1.0\linewidth}{!}{
  \begin{tabular}{lcccccccc}
    \toprule
    \multirow{2}{*}{Models} & \multirow{2}{*}{Methods} & \multicolumn{3}{c}{ALFWorld} & \multicolumn{3}{c}{PDDL}\\
    \cmidrule(lr){3-5} \cmidrule(lr){6-8}
    & &$At$(M) & $Ft$ & $Ct$(M) & $At$(M) & $Ft$ & $Ct$(M) \\
    \cmidrule(lr){1-8}
    \multirow{4}{*}{Qwen3-32B}
    & ReasoningBank       & 0.04 & 0.74&0.05&\textbf{0.07} & 0.19&0.34\\
    & G-Memory                & 0.05& 0.93&0.06 & 0.12&0.21& 0.59\\
    & Hiplan       & 0.09 & 0.96&0.10&0.21 &0.18 & 1.17\\
    & \textbf{M-Plan}        & \textbf{0.03} & \textbf{1.00} &\textbf{0.03}& 0.17& \textbf{0.54}&\textbf{0.32} \\
    \cmidrule(lr){1-8}  
    \multirow{4}{*}{Qwen3-14B}
    & ReasoningBank       & 0.04& 0.69 &0.06& \textbf{0.07}&0.13& 0.54\\
    & G-Memory                & 0.05 & 0.89&0.06&0.12&0.20 & 0.63\\
    & Hiplan      & 0.11 & 0.91&0.12&0.21&0.11&1.89\\
    & \textbf{M-Plan}        & \textbf{0.04} & \textbf{0.94}&\textbf{0.05}&0.16 &\textbf{0.45} &\textbf{0.36} \\
    \bottomrule
  \end{tabular}
  }
\end{table}

\begin{table}
  \caption{Ablation study of M-Plan on PDDL with Qwen3-32B. \textit{Retrieval} indicates whether the short-term memory module receives cases from long-term memory. All \texttt{SR.} and $AD$ values are reported in percentage (\%).}
  \label{ablation}
  \centering
  \small
\resizebox{1.0\linewidth}{!}{
  \begin{tabular}{lcccccccc}
    \toprule
    \multirow{2}{*}{Methods} 
    & \multicolumn{4}{c}{Modules} 
    & \multicolumn{4}{c}{Performance} \\
    \cmidrule(lr){2-5} \cmidrule(lr){6-9}
    & \textit{Optimizer} & \textit{Retrieval} & \textit{Insighter} & \textit{Hinter}
    & \texttt{SR.} & $AD$ & \texttt{AS.} & \texttt{Meta-Step.}\\
    \cmidrule(lr){1-9}
    Standard     
    & \cmark & \cmark & \cmark & \cmark 
    & \textbf{52.8\textpm0.9} & \textbf{74.3\textpm1.3} & \textbf{22.30} &\textbf{29.22}\\
    
    w/o \textit{Optimizer}  
    & \xmark & \cmark & \cmark & \cmark 
    &47.8\textpm2.4&68.2\textpm2.0&22.35&29.58\\
    
    w/o \textit{Retrieval}   
    & \cmark & \xmark & \cmark & \cmark 
    &42.5\textpm0.8&61.4\textpm0.9&23.63&35.12\\
    
    w/o \textit{Insighter}   
    & \cmark & \cmark & \xmark & \cmark 
    &48.3\textpm2.9&66.8\textpm1.5&22.81&31.44\\
    
    w/o \textit{Hinter}  
    & \cmark & \cmark & \cmark & \xmark 
    &41.1\textpm1.9&61.6\textpm1.8&24.01&37.64\\
    \bottomrule
  \end{tabular}
  }
\end{table}

\subsection{Ablation study}

To further demonstrate the effectiveness and efficiency of each module in M-Plan, we conduct ablation experiments on PDDL tasks using the Qwen3-32B model, with results summarized in Table~\ref{ablation}. Compared with the full standard model, removing the \textit{Optimizer} reduces \texttt{SR.} from 52.8 to 47.8 and lowers $AD$ from 74.3 to 68.2, while \texttt{Meta-Step.} slightly rises. This confirms that the \textit{Optimizer} helps refine generated plans and mitigates flawed reasoning, lifting the overall success rate and plan quality.

When the \textit{Retrieval} module is disabled so that the short-term memory cannot access historical cases stored in long-term memory, we observe a substantial performance degradation: \texttt{SR.} falls to 42.5, $AD$ drops to 61.4, and both \texttt{AS.} and \texttt{Meta-Step.} increase markedly. This verifies that retrieved prior planning cases supply useful experience for the \textit{Planner} and \textit{Insighter}, reducing redundant exploration and supporting more efficient reasoning.

Ablating the \textit{Insighter} also harms performance. The variant without \textit{Insighter} achieves \texttt{SR.}=48.3 and $AD$=66.8, accompanied by higher \texttt{AS.} and \texttt{Meta-Step.}. This indicates that task-specific insights extracted by the \textit{Insighter} offer complementary cues that regular planning modules lack, helping the model better capture task constraints and produce more reasonable plans. Finally, removing the \textit{Hinter} yields the lowest \texttt{SR.} (41.1) and $AD$ (61.6), together with the largest \texttt{Meta-Step.} of all ablated variants. Such a severe regression highlights the critical role of the \textit{Hinter}: it delivers intermediate hints during iterative planning, guides the model away from dead-end paths, and suppresses excessive useless reasoning steps.

Across all variants, every proposed module contributes positively to the full M-Plan pipeline. Their combined effect improves task success, enhances plan quality measured by $AD$, and reduces redundant reasoning overhead in PDDL planning.

\section{Conclusion}
\label{sec:conclusion}
In this paper, we have proposed a comprehensive planning framework for LLM agents with memory arrangement. M-Plan combines long short-term memory and global-local guidance together, which can boost LLM agents to tackle multi-turn or other similar complex tasks. The framework can address the limitations of reusing, high cost, not fully guided by abundant global and conditional local guidance, and lacking evolution mechanisms by long short-term memory coordination with closed-loop optimization. The experiments on simulated environment benchmarks, ALFWorld, Webshop and PDDL fully show more admirable effectiveness and efficiency than baselines. 

\textbf{Limitation:} While M-Plan is mature in long-horizon text processing, there is great potential for M-Plan to be extended to deal with real-world environments, and it can be applied with some compression and storage techniques to enhance engineering value.

\section*{Acknowledgements}
This work is supported by the National Key R\&D Program of China (No. 2026YFE0101200), and the National Natural Science Foundation of China (No. 62671175). The computations in this research were performed using the CFFF platform of Fudan University.

\bibliographystyle{unsrt}
\bibliography{reference}

\appendix
\section{PROMPT LIBRARY}
\label{A}
In each prompt, the action space comes from the benchmark official set, and one or two simple written templates can be applied to further constrain the output of the corresponding agent.

\subsection{ACTIONER}
\includegraphics[width=0.9\linewidth]{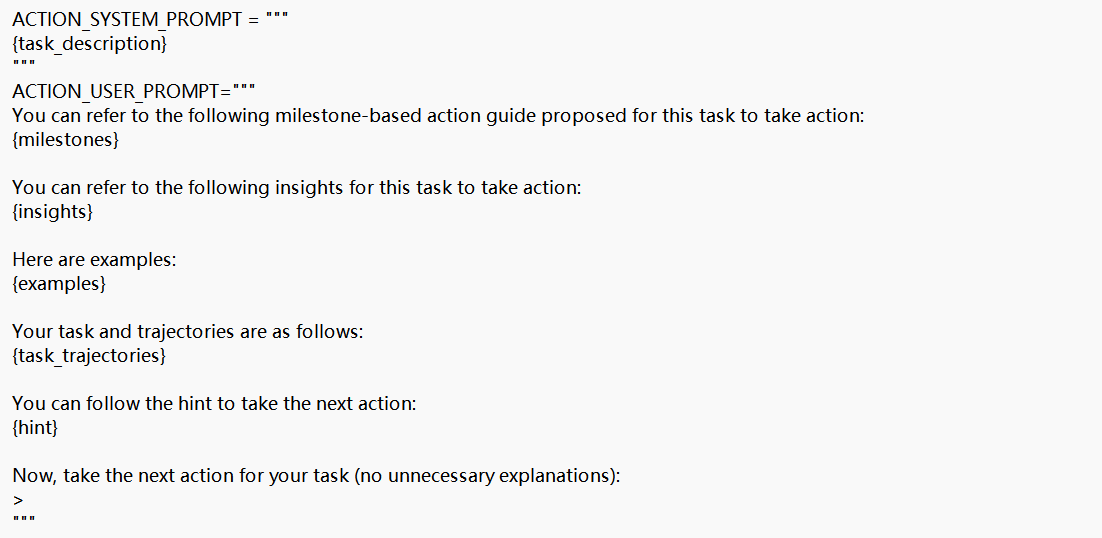}
\vspace{1em} 

\subsection{PLANNER}
\includegraphics[width=0.9\linewidth]{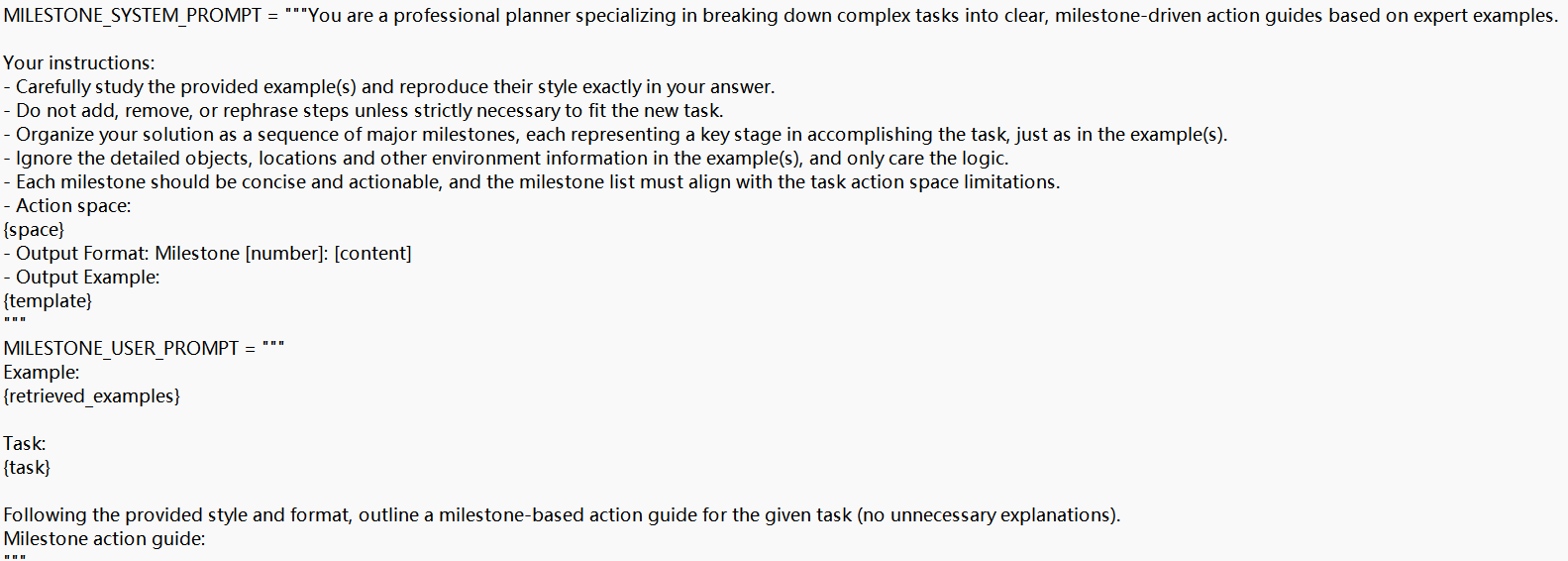}
\vspace{1em}

\subsection{INSIGHTER}
\includegraphics[width=0.9\linewidth]{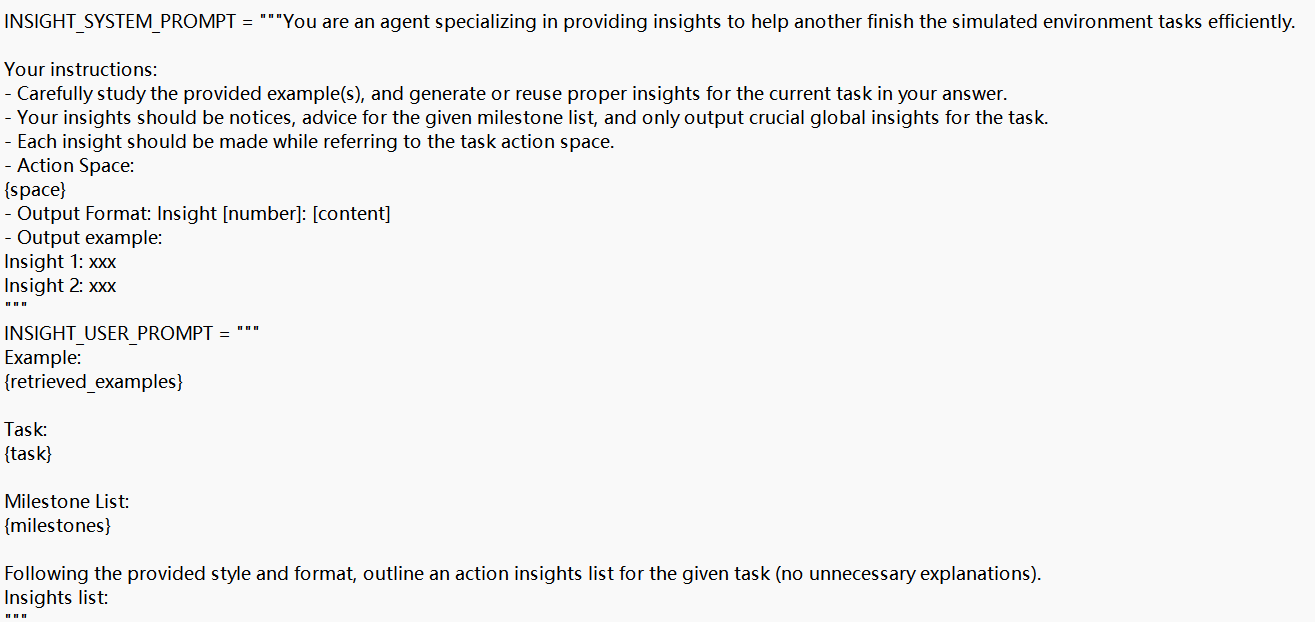}
\vspace{1em}

\subsection{HINTER}
\includegraphics[width=0.9\linewidth]{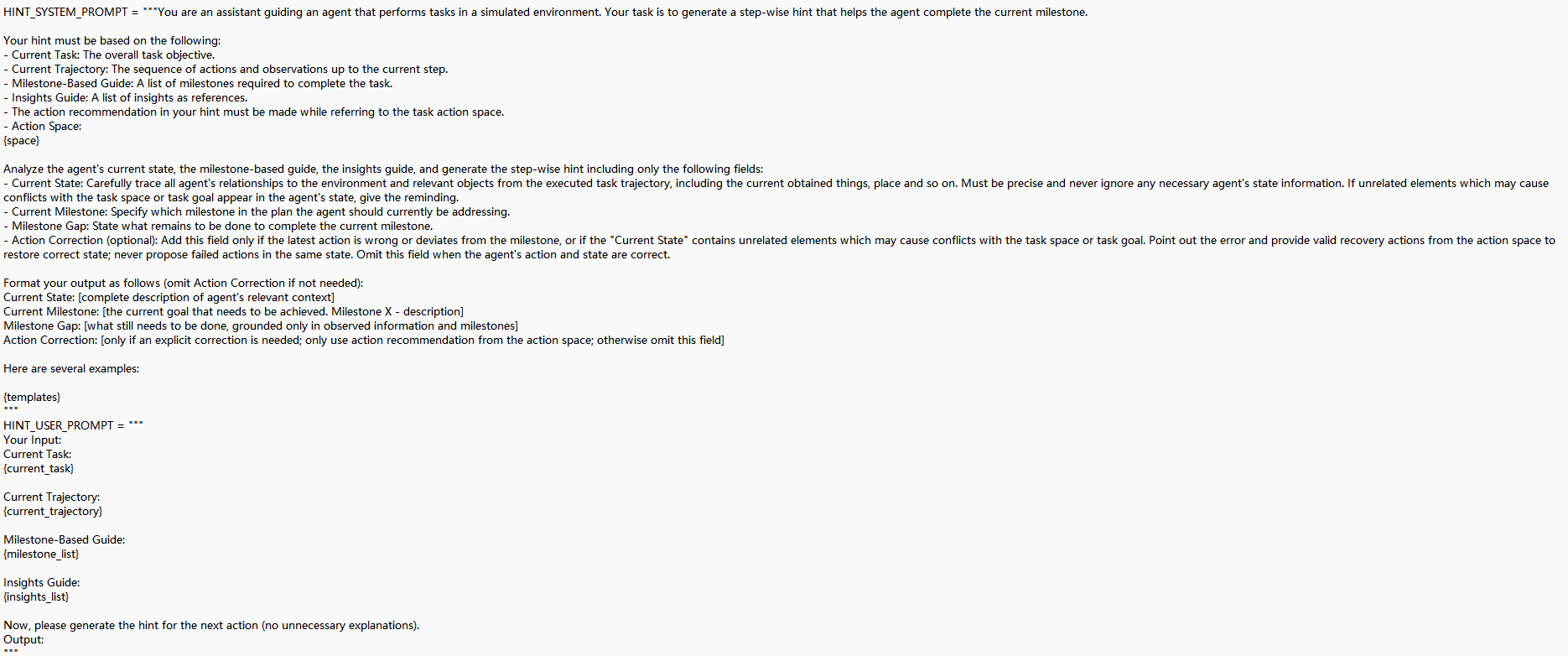}
\vspace{1em}

\subsection{SUCCESS OPTIMIZER}
\includegraphics[width=0.9\linewidth]{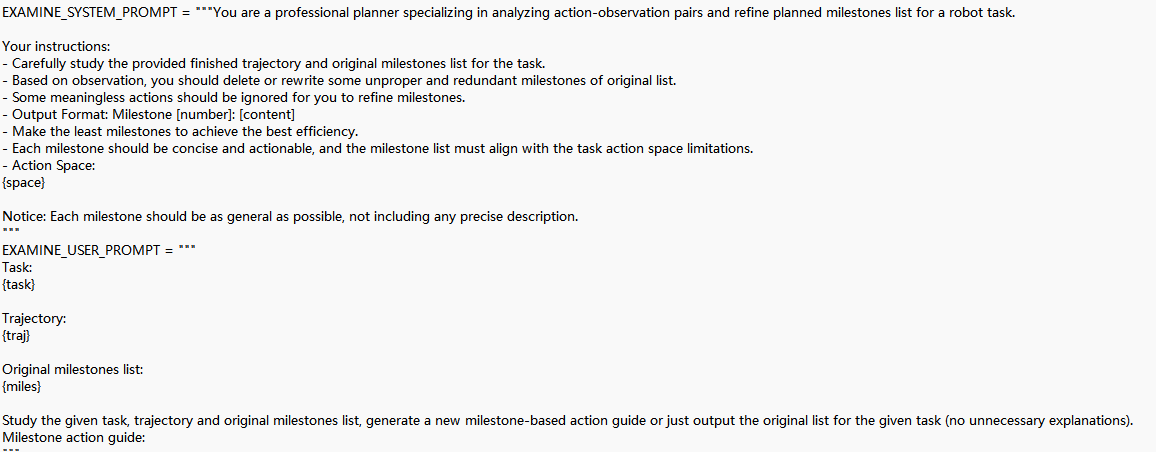}
\vspace{1em}

\subsection{FAILED OPTIMIZER}
\includegraphics[width=0.9\linewidth]{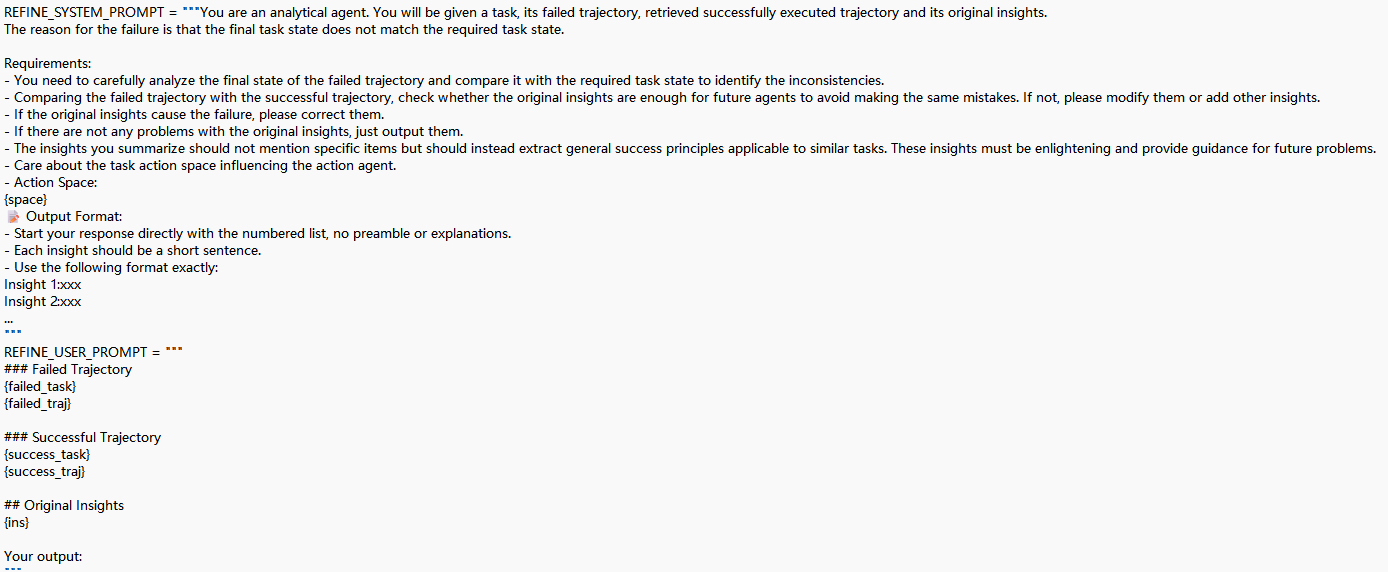}

\section{FORMULA REASONING}
\label{B}
First, we assume that in a benchmark, the average reward(profit) one successful task can bring is $\alpha$, and the average punishment(loss) is $\beta$, which is larger than 0 in contrast to $\alpha$. To avoid that any of them appears in the formula individually and to counterbalance them, we can write an original division form to express the benefit of the success over the failure :
\begin{equation}
    Pt'=\frac{\alpha SR}{\beta FR},\quad (0\le SR<1,\ \beta>0),
\end{equation}
\begin{equation}
    FR=1-SR.
\end{equation}
Since $\frac{\alpha}{\beta}$ is a constant for a fixed benchmark, we can simplify the expression to obtain:
\begin{equation}
    Pt=\frac{SR}{FR}.
\end{equation}
However, this direct division suffers from an edge-explosion problem as $SR$ approaches 1. To constrain unbounded growth and normalize the upper bound, we introduce the $acrtan$ function:
\begin{equation}
    Ft=\frac{2}{\pi}\arctan(Pt),\quad
    Ct=\frac{At}{Ft}.
\end{equation}
Treating $SR$ as the independent variable $x$, we compute the first- and second-order derivatives of $Ft$ with respect to $x$:
\begin{equation}
    \frac{\mathrm{d}Ft}{\mathrm{d}x} = \frac{2}{\pi\big(x^2+(1-x)^2\big)},
\end{equation}
\begin{equation}
    \frac{\mathrm{d}^2 Ft}{\mathrm{d}x^2} = \frac{4(1-2x)}{\pi\big(x^2+(1-x)^2\big)^2}.
\end{equation}
The second-order derivative reveals that $Ft$ attains its maximum at $SR=0.5$. Specifically, the first-order derivative increases when $SR<0.5$ and decreases when $SR>0.5$. This property encourages growth speed for low values of $SR$ below 0.5 and mitigates the rate of increase speed for larger $SR$.

\section{EXPERIMENT SETUP}
\label{C}

\textbf{Datasets} We evaluate M-Plan on three widely test multi-turn benchmarks for long-horizon planning and decision-making:
\begin{enumerate}[leftmargin=2em, itemsep=2pt, parsep=0pt, topsep=0pt, partopsep=0pt]
    \item \textbf{ALFWorld} \cite{shridhar2021alfworldaligningtextembodied} is a simulated household environment base on text interaction. This benchmark comprises six kinds of subsets: put, put two, examine under the light, clean \& put, heat \& put, and cool \& put, checking the capacity of an agent for understanding its own state and changes of environment via interactions. The maximum interaction step this benchmark can proceed is 50, also taken in our experiments. And we conduct tests on the officially established split of  134 out-of-distribution tasks.
    \item \textbf{Webshop} \cite{yao2023webshopscalablerealworldweb} is a benchmark simulating an online shopping website, filled with over 1.18 million products. It will require the agent to search for target commodities and click some buttons to buy the ideal product, while reading different web pages. We conduct experiments on a set of 200 instructions. We set the max steps as 40.
    \item \textbf{PDDL} \cite{ma2024agentboardanalyticalevaluationboard} is a benchmark simulating four kinds of robot tasks, including making cocktails, placing blocks, moving balls and dealing with tyres. The agent is required to complete the specified goal, which typically comprises multiple subgoals within a 30-step budget.
\end{enumerate}

\textbf{Baselines}: we compare M-Plan with the following LLM-based agent baselines, ranging from classic short-term memory reasoning to long-term memory retrieval paradigms, including:

\begin{enumerate}[leftmargin=2em, itemsep=2pt, parsep=0pt, topsep=0pt, partopsep=0pt]
    \item \textbf{G-Memory} \cite{zhang2025gmemorytracinghierarchicalmemory} is a representative retrieval-augmented memory approach to reuse detailed trajectories, plan-like key steps, and history summarized insights, and places many agents to refine or integrate insights for both successful and failed task trajectories.
    \item \textbf{ReasoningBank} \cite{ouyang2026reasoningbankscalingagentselfevolving} is a typical extraction and reusing long-term memory method. The retrieved memory item is a series of collections of insights, comprising titles, descriptions, and content of each title, summarized by an agent, used to guide similar tasks.
    \item \textbf{Hiplan} \cite{li2025hiplanhierarchicalplanningllmbased} is a short-term memory method equipped with a prepared memory library, containing a few trajectories to be retrieved. It also uses milestones as global guidance and step-wise hints. During the execution, similar milestones and trajectory segments will be searched to generate a hint in each step.
\end{enumerate}

\textbf{Reproduction Details}. Since no official trajectory data are provided for Hiplan~\cite{li2025hiplanhierarchicalplanningllmbased}, we carefully construct six expert trajectories for ALFWorld, each corresponding to one of its six task types, as well as 25 task trajectories for Webshop. Following the paper, all trajectories are annotated with milestones and segments. Rather than providing an excessive number of cases, we aim to use a compact set of high-quality trajectories to improve performance. All preparation activities are completed using Qwen3-32B. For long-term memory methods and M-Plan, to ensure a fair alignment between our method and the compared approaches, we do not provide any prepared trajectory shots. The models used in the experiments are all deployed on two Nvidia A100 GPUs.

\section{HINTER DETAIL}
\label{D}
Precisely, we establish two distinct principles tailored for different scenarios: 
\begin{enumerate}
    \item Loose Triggering: In ALFWorld and Webshop, agents often perform redundant revisiting and rechecking actions, which may trap them in persistent loops. Accordingly, for such occasions, once repeated actions appear in the action history, hints will be generated to provide guidance, as such repetitive behaviors tend to induce further troubles in these environments. Hints are also triggered if the most recent action is invalid.
    \item Strict Triggering: In PDDL environments, hint activation follows a stricter rule: hints are triggered only when the current action duplicates the immediately preceding action or is invalid. This design permits permissible repeated actions, preserves task-solving effectiveness, and improves efficiency.
\end{enumerate}
Additionally, since the environment observation can directly indicate whether the current action is valid or invalid, we use this feedback for action validation without introducing an additional agent, thereby further reducing system complexity and computational overhead. However, another issue arises with the $Hinter$ module. Although it is designed to help the $Actioner$ escape potential traps, the $Hinter$ itself may produce repetitive unhelpful or misleading guidance. For this reason, we introduce a mechanism to suspend the $Hinter$ when the next action would trigger the same hint repeatedly. After getting paused, the $Hinter$ awaits further information from the $Actioner$ to think again.

\section{STD RESULTS}
\label{E}
\begin{table}[htbp!]
  \caption{Comparison of average steps on ALFWorld and Webshop. We report \texttt{AS.}, \texttt{ASS.}, and \texttt{Meta-Step}, while \texttt{AS.} is not considered for Webshop due to its specific characteristics, as discussed in Sec.~\ref{metric}.}
  \label{steps_std}
  \centering
  \setlength{\tabcolsep}{2pt}
  \begin{tabular}{l c c c c c c}
    \toprule
    \multirow{2}{*}{Models} & \multirow{2}{*}{Methods} & \multicolumn{3}{c}{ALFWorld} & \multicolumn{2}{c}{Webshop}\\
    \cmidrule(lr){3-5} \cmidrule(lr){6-7}
    & & \texttt{AS.} & \texttt{ASS.} & \texttt{Meta-Step} & \texttt{ASS.} & \texttt{Meta-Step} \\
    \cmidrule(lr){1-7}
    \multirow{4}{*}{Qwen3-32B}
    & ReasoningBank       & 24.99±0.00 & 13.96±0.00 & 20.12±0.00& 6.14±0.02 & 22.90±0.92\\
    & G-Memory            & 24.93±0.48 & 20.53±0.51 & 24.14±0.60 & \textbf{5.10±0.24} & 10.90±0.69\\
    & Hiplan              & 18.38±0.36 & 16.46±0.08 & 17.46±0.25 & 8.09±0.00 & 20.48±0.00\\
    & \textbf{M-Plan}     & \textbf{12.48±0.00} & \textbf{12.20±0.00} & \textbf{12.28±0.00} & 5.35±0.00 & \textbf{9.47±0.00}\\
    \cmidrule(lr){1-7}
    \multirow{4}{*}{Qwen3-14B}
    & ReasoningBank       & 26.30±0.00 & 13.91±0.00 & 21.17±0.00 & 5.90±0.13 & 20.93±0.02\\
    & G-Memory            & 18.66±0.30 & 12.94±0.16 & 15.30±0.27& \textbf{5.13±0.08} & 8.96±0.31\\
    & Hiplan              & 20.40±0.31 & 16.28±0.32 & 18.55±0.01 & 8.05±0.00 & 25.16±0.00 \\
    & \textbf{M-Plan}     & \textbf{14.63±0.04} & \textbf{11.05±0.19} & \textbf{12.17±0.16} & 5.48±0.06 & \textbf{8.03±0.06}\\
    \bottomrule
  \end{tabular}
\end{table}

\begin{table}[htbp!]
  \caption{Comparison of average steps on PDDL. We report \texttt{AS.}, \texttt{ASS.}, and \texttt{Meta-Step}.}
  \label{steps_std}
  \centering
  \setlength{\tabcolsep}{2pt}
  \begin{tabular}{l c c c c}
    \toprule
    \multirow{2}{*}{Models} & \multirow{2}{*}{Methods} & \multicolumn{3}{c}{PDDL}\\
    \cmidrule(lr){3-5}
    & & \texttt{AS.} & \texttt{ASS.} & \texttt{Meta-Step}  \\
    \cmidrule(lr){1-5}
    \multirow{4}{*}{Qwen3-32B}
    & ReasoningBank      &26.61±0.31 &15.76±1.84 & 65.90±0.15\\
    & G-Memory            &26.95±0.69 &18.16±1.90 & 71.73±0.43 \\
    & Hiplan              &26.33±0.14 &\textbf{13.48±0.10} & 60.70±0.07\\
    & \textbf{M-Plan}     &\textbf{22.30±0.13} & 15.42±0.03 & \textbf{29.22±0.01} \\
    \cmidrule(lr){1-5}
    \multirow{4}{*}{Qwen3-14B}
    & ReasoningBank       &28.16±0.16 &19.33±0.30 & 112.42±0.25\\
    & G-Memory            1 &26.97±0.07 &17.54±1.15 & 71.86±0.05\\
    & Hiplan               &27.56±0.28 &\textbf{13.71±0.61} &92.18±0.38 \\
    & \textbf{M-Plan}     &\textbf{23.33±0.33} &15.54±0.11 & \textbf{33.75±0.05}\\
    \bottomrule
  \end{tabular}
\end{table}

\begin{table}[t]
  \caption{Token-efficiency comparison. We report $At$, $Ft$ and $Ct$ (see in Sec.~\ref{metric}), mainly focusing on the comparison of $Ct$. Webshop is not discussed for the fairness in Sec.~\ref{metric}.}
  \label{token_std}
  \centering
  \resizebox{1.0\linewidth}{!}{
  \begin{tabular}{lcccccccc}
    \toprule
    \multirow{2}{*}{Models} & \multirow{2}{*}{Methods} & \multicolumn{3}{c}{ALFWorld} & \multicolumn{3}{c}{PDDL}\\
    \cmidrule(lr){3-5} \cmidrule(lr){6-8}
    & &$At$(M) & $Ft$ & $Ct$(M) & $At$(M) & $Ft$ & $Ct$(M) \\
    \cmidrule(lr){1-8}
    \multirow{4}{*}{Qwen3-32B}
    & ReasoningBank       & 0.04±0.00 & 0.74±0.00&0.05±0.00&\textbf{0.07±0.00} & 0.19±0.01&0.34±0.01 \\
    & G-Memory                & 0.05±0.00 & 0.93±0.01&0.06±0.00 & 0.12±0.00&0.21±0.02 & 0.59±0.07\\
    & Hiplan       & 0.09±0.00 & 0.96±0.01&0.10±0.00 &0.21±0.00 &0.18±0.01 & 1.17±0.06\\
    & \textbf{M-Plan}        & \textbf{0.03±0.00} & \textbf{1.00±0.00} &\textbf{0.03±0.00}& 0.17±0.00& \textbf{0.54±0.01}&\textbf{0.32±0.01} \\
    \cmidrule(lr){1-8}  
    \multirow{4}{*}{Qwen3-14B}
    & ReasoningBank       & 0.04±0.00& 0.69±0.00 &0.06±0.00& \textbf{0.07±0.00}&0.13±0.01 & 0.54±0.03\\
    & G-Memory                & 0.05±0.00 & 0.89±0.01&0.06±0.00&0.12±0.01 &0.20±0.02 & 0.63±0.04\\
    & Hiplan      & 0.11±0.00 & 0.91±0.01&0.12±0.00&0.21±0.00 &0.11±0.01 &1.89±0.26 \\
    & \textbf{M-Plan}        & \textbf{0.04±0.00} & \textbf{0.94±0.00}&\textbf{0.05±0.00}&0.16±0.01 &\textbf{0.45±0.02} &\textbf{0.36±0.00} \\
    \bottomrule
  \end{tabular}
  }
\end{table}

\begin{table}[t]
  \caption{Ablation study of M-Plan on PDDL with Qwen3-32B. \textit{Retrieval} indicates whether the short-term memory module receives cases from long-term memory. All \texttt{SR.} and $AD$ values are reported in percentage (\%).}
  \label{ablation_std}
  \centering
  \small
\resizebox{1.0\linewidth}{!}{
  \begin{tabular}{lcccccccc}
    \toprule
    \multirow{2}{*}{Methods} 
    & \multicolumn{4}{c}{Modules} 
    & \multicolumn{4}{c}{Performance} \\
    \cmidrule(lr){2-5} \cmidrule(lr){6-9}
    & \textit{Optimizer} & \textit{Retrieval} & \textit{Insighter} & \textit{Hinter}
    & \texttt{SR.} & $AD$ & \texttt{AS.} & \texttt{Meta-Step.}\\
    \cmidrule(lr){1-9}
    Standard     
    & \cmark & \cmark & \cmark & \cmark 
    & \textbf{52.8\textpm0.9} & \textbf{74.3\textpm1.3} & \textbf{22.30\textpm0.13} &\textbf{29.22\textpm0.01}\\
    
    w/o \textit{Optimizer}  
    & \xmark & \cmark & \cmark & \cmark 
    &47.8\textpm2.4&68.2\textpm2.0&22.35\textpm0.04&29.58\textpm0.02\\
    
    w/o \textit{Retrieval}   
    & \cmark & \xmark & \cmark & \cmark 
    &42.5\textpm0.8&61.4\textpm0.9&23.63\textpm0.07&35.12\textpm0.03\\
    
    w/o \textit{Insighter}   
    & \cmark & \cmark & \xmark & \cmark 
    &48.3\textpm2.9&66.8\textpm1.5&22.81\textpm0.59&31.44\textpm0.08\\
    
    w/o \textit{Hinter}  
    & \cmark & \cmark & \cmark & \xmark 
    &41.1\textpm1.9&61.6\textpm1.8&24.01\textpm0.42&37.64\textpm0.08\\
    \bottomrule
  \end{tabular}
  }
\end{table}

\end{document}